\documentclass[letterpaper, 10 pt, conference]{ieeeconf}  % Comment this line out if you need a4paper

\IEEEoverridecommandlockouts                              % This command is only needed if 
\title{\LARGE \bf
FARE: Fast-Slow Agentic Robotic Exploration
}

\author{Shuhao Liao$^{1,2}$, Xuxin Lv$^{1}$, Jeric Lew$^{2}$, Shizhe Zhang$^{2}$, Jingsong Liang$^{2}$,  Peizhuo Li$^{2}$, \\Yuhong Cao$^{2}$, Wenjun Wu$^{1}$, Guillaume Sartoretti$^{2}$%
\thanks{$^{1}$Beihang University, China}
\thanks{$^{2}$Department of Mechanical Engineering, National University of Singapore, Singapore}
}

\usepackage{amssymb}
\usepackage{amsmath}
\usepackage{graphicx}
\usepackage{multirow}
\usepackage{booktabs}
\usepackage{tabularx}
\usepackage{subcaption}
\usepackage{cite}
\usepackage{tcolorbox}
\usepackage[ruled]{algorithm2e}

\begin{document}

\makeatletter
\let\@oldmaketitle\@maketitle% Store \@maketitle
\renewcommand{\@maketitle}{\@oldmaketitle% Update \@maketitle to insert...
% \vspace{-.45cm}
  \begin{center}
  \vspace{1.0em}
  \captionsetup{type=figure}
  \includegraphics[width=1\textwidth]{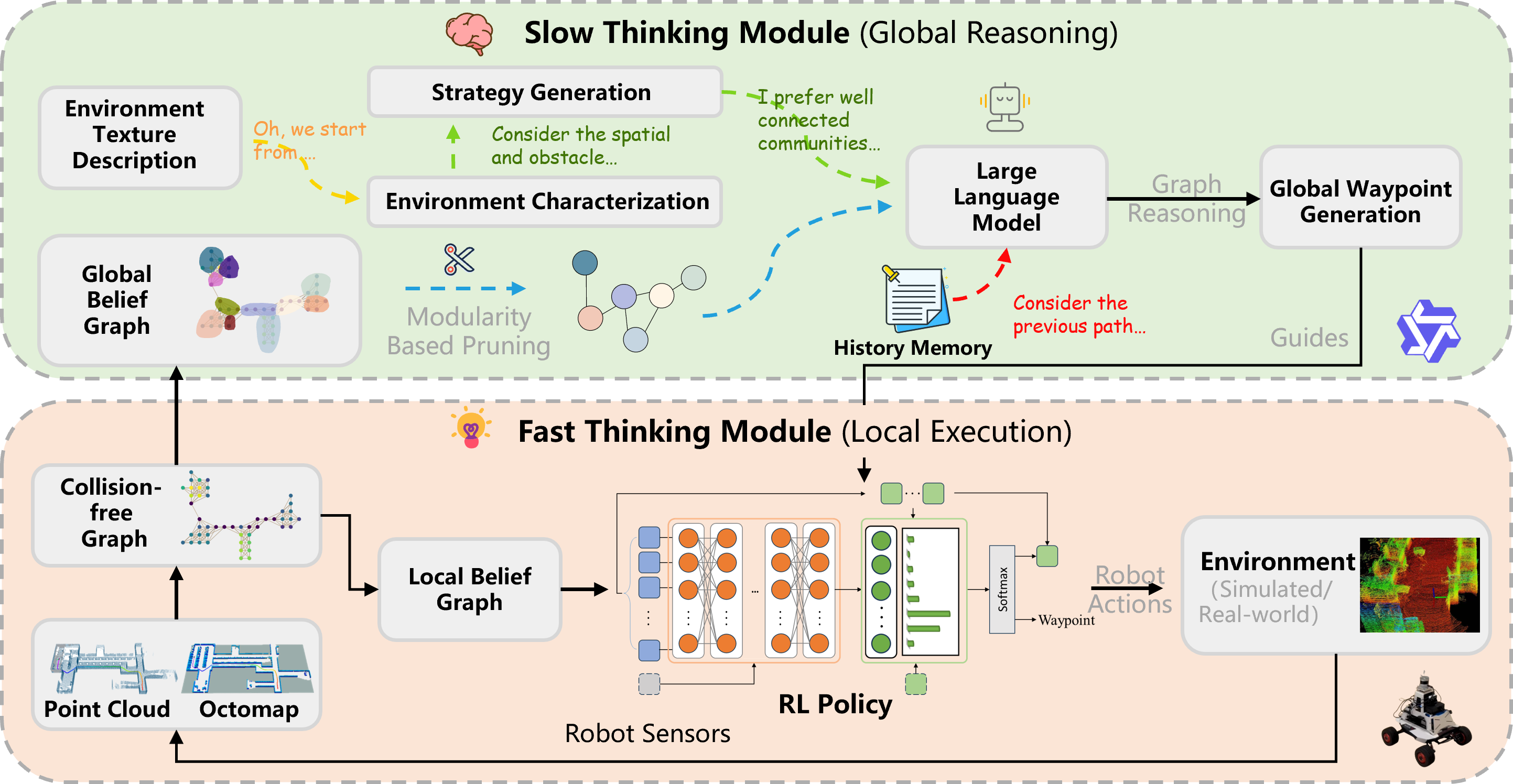}
    \captionof{figure}{Overall framework of FARE, where a slow-thinking LLM module performs global reasoning, and a fast-thinking RL module executes locally grounded, sensor-driven autonomous exploration under this global guidance.} 
    \vspace{-1.0em}
    \label{pic1}
    \setcounter{figure}{1} % 把 figure 编号设为1 → 下一张图会变成 Fig. 2
  \end{center}
}
\makeatother

\maketitle
  
\thispagestyle{empty}
\pagestyle{empty}

%%%%%%%%%%%%%%%%%%%%%%%%%%%%%%%%%%%%%%%%%%%%%%%%%%%%%%%%%%%%%%%%%%%%%%%%%%%%%%%%
%%%%%%%%%%%%%%%%%%%%%%%%%%%%%%%%%%%%%%%%%%%%%%%%%%%%%%%%%%%%%%%%%%%%%%%%%%%%%%%%

\begin{abstract}

This work advances autonomous robot exploration by integrating agent-level semantic reasoning with fast local control. We introduce FARE, a hierarchical autonomous exploration framework that integrates a large language model (LLM) for global reasoning with a reinforcement learning (RL) policy for local decision making. FARE follows a fast-slow thinking paradigm. The slow-thinking LLM module interprets a concise textual description of the unknown environment and synthesizes an agent-level exploration strategy, which is then grounded into a sequence of global waypoints through a topological graph. To further improve reasoning efficiency, this module employs a modularity-based pruning mechanism that reduces redundant graph structures. The fast-thinking RL module executes exploration by reacting to local observations while being guided by the LLM-generated global waypoints. The RL policy is additionally shaped by a reward term that encourages adherence to the global waypoints, enabling coherent and robust closed-loop behavior. This architecture decouples semantic reasoning from geometric decision, allowing each module to operate in its appropriate temporal and spatial scale. In challenging simulated environments, our results show that FARE achieves substantial improvements in exploration efficiency over state-of-the-art baselines. We further deploy FARE on hardware and validate it in complex, large scale $200m\times130m$ building environment.

\end{abstract}

%%%%%%%%%%%%%%%%%%%%%%%%%%%%%%%%%%%%%%%%%%%%%%%%%%%%%%%%%%%%%%%%%%%%%%%%%%%%%%%%
%%%%%%%%%%%%%%%%%%%%%%%%%%%%%%%%%%%%%%%%%%%%%%%%%%%%%%%%%%%%%%%%%%%%%%%%%%%%%%%%

\section{INTRODUCTION}

Autonomous exploration is a core capability for mobile robots operating in unknown environments, where the objective is to efficiently acquire geometric information. The robot acquires environmental data through sensors such as 3D LiDARs or cameras and incrementally builds a representation of the environment, commonly encoded as an occupancy grid or voxel map.
With the reliability of modern LiDAR odometry and SLAM (Simultaneous Localization and Mapping), recent work has increasingly focused on the exploration problem itself, without concerns about mapping or localization accuracy~\cite{selin2019efficient,cao2021tare,dang2020graph,peltzer2022fig,huang2023fael,long2024hphs,cao2025header}. Yet, despite substantial progress, efficient exploration in complex real-world environments remains challenging. Existing planners struggle to leverage long-term structural information embedded in the evolving map and often rely on fixed heuristics or rigid decision rules that limit their ability to adapt exploration strategies to different environmental characteristics. 

Existing autonomous exploration approaches are largely divided into conventional methods and learning based methods. Conventional planners maintain a geometric belief map and plan motions that trade off investigating new regions and further refining partially explored areas~\cite{yamauchi1997frontier,bircher2016receding,selin2019efficient,dang2020graph}. To improve scalability in large and complex environments, recent conventional planners increasingly rely on hierarchical planning, which decouples exploration into a low-resolution global path computed over a coarse belief and a high-resolution local path computed over the nearby detailed map~\cite{zhu2021dsvp,cao2021tare,huang2023fael,long2024hphs}. The key insight there is that fine reasoning is most effective locally, while coarse inference suffices for guiding long-term exploration. By contrast, learning-based methods formulate exploration as a partially observable Markov decision process and use RL to learn policies that map observations to next waypoint~\cite{chen2019self,chen2020autonomous,ariadne,cao2024deep}. These methods improve exploration efficiency through reward designs that encourage informative observations and reduce redundant motions, often supported by neural architectures that capture spatial correlations in the evolving map. 

Despite these advances, existing methods still struggle to exploit long-term information and to adapt their exploration strategy in response to evolving environments. 
Conventional planners rely on fixed hyperparameters that determine how they balance path length with information gain or how they choose the resolution of the global representation. These parameters typically remain constant throughout deployment, which prevents the planner from adjusting its behavior in different environments. As a result, the planner may act too conservatively in open spaces or spend unnecessary effort refining cluttered areas. Even hierarchical planners face similar constraints. Their performance depends on a predefined coarse-to-fine decomposition of the map, which restricts how global structural information can influence local decisions.
Learning-based planners encounter a different but equally important limitation. The objective in autonomous exploration is to minimize the overall time or distance required to complete coverage, yet this objective only becomes observable at the end of an episode. The reward signal therefore becomes extremely sparse, making it difficult for the policy to learn behaviors that depend on long-term consequences. Existing methods address this issue by introducing dense rewards that provide immediate feedback proportional to newly observed information~\cite{chen2019self,chen2020autonomous,xu2022explore,zhu2024maexp}. Although this improves training stability, it also biases the optimization process toward short-term, immediately observable information gain.
As a result, long-horizon credit assignment becomes significantly more challenging in practice, making it difficult for the policy to reliably acquire behaviors such as long-distance backtracking or exploiting distant structural cues that only become relevant later in the mission. 
Consequently, both conventional and learning-based approaches still struggle to utilize long-term information effectively and to adjust their exploration strategy as environmental conditions change.

Inspired by the impressive intelligence emerging in recent AI developments, we propose FARE, a novel hierarchical fast–slow thinking framework that separates global reasoning from local decision making. 
Our slow-thinking module operates at the global level and reasons over a sparse belief graph derived from community detection and modularity-based clustering~\cite{cao2025header}. Given a concise textual description of the environment, our slow-thinking module extracts high-level spatial characteristics and obstacle configurations, which are then used to synthesize an exploration strategy tailored to the overall map structure. This strategy is grounded onto the global belief graph to generate a sequence of global waypoints that reflect long-range coverage priorities and adapt as the partial map is incrementally updated.
Our fast-thinking module operates at the local level and uses a reinforcement learning policy to select actions based on dense local map information, including frontiers and nearby geometric features. During deployment, the policy reacts in real time to local observations while also incorporating the global waypoints supplied by the slow-thinking module. To ensure that the learned policy can follow long-horizon guidance without relying on overly myopic reward shaping, the training process includes an additional reward term that encourages consistency with the global waypoints. This design allows the policy to remain responsive to local conditions while still pursuing long-term objectives.
Through this structured interaction, our framework effectively integrates global reasoning with local decision, enabling the robot to exploit long-term information and to adapt its exploration behavior as the environment unfolds. 
As demonstrated in simulations and real-world robot validation, this design reduces unnecessary backtracking and yields more efficient exploration trajectories.

%%%%%%%%%%%%%%%%%%%%%%%%%%%%%%%%%%%%%%%%%%%%%%%%%%%%%%%%%%%%%%%%%%%%%%%%%%%%%%%%
%%%%%%%%%%%%%%%%%%%%%%%%%%%%%%%%%%%%%%%%%%%%%%%%%%%%%%%%%%%%%%%%%%%%%%%%%%%%%%%%

\section{RELATED WORK}

\subsection{Autonomous Exploration}
Autonomous exploration has been widely studied using both conventional and learning-based approaches. 
Conventional planners are predominantly frontier-based, selecting exploration targets at the boundary between known and unknown regions, where early methods directly navigate to individual frontier points, leading to greedy and myopic behavior~\cite{gonzalez2002navigation}. 
Later works reformulate exploration as a viewpoint selection problem, in which candidate viewpoints are sampled and evaluated based on expected information gain and feasibility, enabling non-myopic planning through sampling-based strategies~\cite{bircher2016receding,xu2021autonomous}. 
To improve scalability in large environments, many works adopt hierarchical designs that combine fine-grained local planning with coarse global representations. 
Representative approaches construct sparse global frontier graphs or viewpoint graphs to guide long-range navigation, while local planners handle detailed viewpoint sampling and collision checking~\cite{cao2021tare,huang2023fael}.
While effective, these methods rely on manually designed global representations and planning heuristics, and require careful balancing between representation sparsity and coverage, which is difficult to achieve in environments with unknown or evolving structures.
Learning-based exploration methods aim to alleviate manual design by training policies to guide exploration decisions. 
Some works employ supervised learning to explicitly predict occupancy or semantic maps from partial observations, and then integrate the predicted maps into conventional frontier-based planners~\cite{chen2019self,xu2022explore}. 
More recent approaches formulate exploration as a sequential decision-making problem and use reinforcement learning to estimate long-term returns from observations. 
Many of these methods rely on convolutional neural networks (CNNs) to process fixed-size map representations or local occupancy grids~\cite{niroui2019deep,chen2019self,xu2022explore,zhu2024maexp}, with different action abstractions ranging from frontier or viewpoint selection~\cite{chen2019self,xu2022explore} to direct navigation commands such as velocity or steering control~\cite{zhu2024maexp,chen2020autonomous}. 
To address the limited scalability and robustness of CNN-based planners, recent works adopt graph learning-based representations and formulate exploration as a next-viewpoint selection problem on a graph~\cite{ariadne,cao2024deep,cao2025header}, enabling variable-sized environments.
Despite these advances, existing methods still struggle to exploit long-term information and adapt their exploration behavior as the environment evolves, largely because exploration decisions are conditioned on local observations or fixed representations, without explicitly leveraging readily available, high-level environment priors to guide planning.

%%%%%%%%%%%%%%%%%%%%%%%%%%%%%%%%%%%%%%%%%%%%%%%%%%%%%%%%%%%%%%%%%%%%%%%%%%%%%%%%

\subsection{LLM-based Graph Reasoning}
LLM-based graph reasoning has advanced substantially in recent years, as an increasing body of work shows that large language models can exploit structured relational information when graphs are presented in suitable, structured formats. 
A central challenge in this line of research lies in how graph structures are encoded and provided to LLMs in a way that preserves relational semantics while remaining compatible with text-based or embedding-based model interfaces. 
To this end, widely adopted approaches introduce structured graph verbalizers, embedding projectors, or instruction-tuned schemas that convert nodes, edges, and local neighborhoods into sequential representations that LLMs can reliably process~\cite{guo2023gpt4graph,Wang2024InstructGraph,chen2024llaga,chen2024exploring}. 
Building on such representations, another line of work focuses on strengthening LLM reasoning capabilities over graphs. 
Recent methods incorporate guided graph traversal, community-aware pruning, and explicitly structured reasoning steps to facilitate multi-hop inference and long-range dependency modeling on complex graphs~\cite{DBLP:conf/iclr/SunXTW0GNSG24,DBLP:conf/aaai/BestaBKGPGGLNNH24,ftog25}. 
In parallel, hybrid approaches couple LLMs with dedicated graph encoders or multimodal modules, allowing symbolic or geometric graph features to complement language-based reasoning, as exemplified by GraphLLM and multimodal extensions such as GITA~\cite{GraphLLM,GITA}. 
Despite these advances, existing LLM-based graph reasoning methods are evaluated primarily on static, offline benchmarks, and rarely address the challenges posed by dynamic, partially observed, and incrementally evolving graph structures that arise in real-world robotic environments.

%%%%%%%%%%%%%%%%%%%%%%%%%%%%%%%%%%%%%%%%%%%%%%%%%%%%%%%%%%%%%%%%%%%%%%%%%%%%%%%%
%%%%%%%%%%%%%%%%%%%%%%%%%%%%%%%%%%%%%%%%%%%%%%%%%%%%%%%%%%%%%%%%%%%%%%%%%%%%%%%%

\section{PROBLEM FORMULATION}

Let $\mathcal{E} \subset \mathbb{R}^2$ denote a bounded and initially unknown environment, represented by a two-dimensional occupancy grid. During exploration, the robot incrementally constructs a partial map $\mathcal{M}$, which is decomposed into known and unknown regions,

\begin{equation}
\mathcal{M}=\mathcal{M}_{\text {u}} \cup \mathcal{M}_{\text {k}} .
\end{equation}

The known region $\mathcal{M}_{\text {k}}$ is further partitioned into free and occupied space $\mathcal{M}_{\text {k}}=\mathcal{M}_{\text {f}} \cup \mathcal{M}_{\text {o}}$, where $\mathcal{M}_{\text {f}}$ denotes traversable space and $\mathcal{M}_{\text {o}}$ denotes occupied space.
At the beginning of exploration, the environment is completely unknown, i.e., $\mathcal{M}=\mathcal{M}_{\text {u}}$. At each decision step, the robot acquires observations using an omnidirectional LiDAR with sensing range $d_s$, and cells within the sensing range are classified as either free or occupied according to traversability.
The objective of autonomous exploration is to compute a collision-free trajectory that minimizes the total traversal cost while completing the exploration,

\begin{equation}
\tau^*=\underset{\tau \in \mathcal{T}}{\operatorname{argmin}} L(\tau), \quad \text { s.t. } \mathcal{M}_{\text {k}}=\mathcal{M}_g
\end{equation}

where $\mathcal{T}$ denotes the set of feasible trajectories and $L: \tau \rightarrow \mathbb{R}^{+}$maps a trajectory to its total path length. While the ground-truth map $\mathcal{M}_g$ is not available during real-world deployment, it is accessible in simulation and benchmark settings for evaluation. In practice, exploration completion is commonly approximated by the absence of remaining unknown or frontier regions.

%%%%%%%%%%%%%%%%%%%%%%%%%%%%%%%%%%%%%%%%%%%%%%%%%%%%%%%%%%%%%%%%%%%%%%%%%%%%%%%%
%%%%%%%%%%%%%%%%%%%%%%%%%%%%%%%%%%%%%%%%%%%%%%%%%%%%%%%%%%%%%%%%%%%%%%%%%%%%%%%%

\section{METHODOLOGY}

%Our method consists of two complementary modules: a slow thinking module and a fast thinking module. The slow thinking module operates at a low frequency and leverages LLMs to perform high-level reasoning based on a global belief graph, incorporating human preferences into the planning process. In contrast, the fast thinking module runs at a higher frequency and performs rapid, real-time inference using a RL model based on local  observations. 

\subsection{Hierarchical Robot Belief Graph}

In this work, we construct a hierarchical robot belief graph through the community-based method proposed in~\cite{cao2025header}. It represents the neighboring area of the robot as a dense local graph, which will serve as the input for the fast-thinking module, and the distant areas as a sparse global graph, which will serve as the input for the slow-thinking module.

\subsubsection{Local Belief Graph}
The robot belief is represented by a collision-free graph constructed from onboard sensor data. The robot trajectory is defined as a sequence of viewpoints \(\tau = (v_0, v_1, \dots)\), where each waypoint \(v_i \in \mathcal{M}_f\) lies in free space.  

At each decision step \(t\), a set of candidate viewpoints \(V_t = \{v_0, v_1, \dots\}\), with \(v_i = (x_i, y_i) \in \mathcal{M}_f\), is uniformly sampled from the current free space following the approach in TARE. Each viewpoint is connected to its \(k\) nearest neighbors, and edges intersecting occupied or unknown regions are removed, yielding a collision-free graph \(G_t = (V_t, E_t)\).

The robot selects viewpoints sequentially, forming a trajectory \(\tau_i \in V_t\). Each node \(v_i\) is assigned a utility value \(u_i\), defined as the number of observable frontiers within sensor range. A frontier \(f_j\) is observable from \(v_i\) if the line segment \(L(v_i, f_j)\) is collision-free and \(\|f_j - v_i\| \le d_s\). The utility is defined as
\begin{equation}
	\begin{aligned}
			&u_i = |F_{o, i}|, \\
			&\forall f_j \in F_{o, i}, \quad \| f_j - v_i \| \leq d_s, \\
			&L(v_i, f_j) \cap (\mathcal{M} - \mathcal{M}_f) = \emptyset.
		\end{aligned}
\end{equation}
where \(F_{o,i}\) denotes the set of observable frontiers at \(v_i\). Then we introduce a square sliding window $\mathcal{W}_{\text{local}}$ of size $d \times d$, centered at the robot’s current position $v_{\text{cur}}$, and construct the \emph{local belief graph} $G_{\text{local}} = (V_{\text{local}}, E_{\text{local}})$. Here, $V_{\text{local}} \subset V_t$ contains all candidate nodes within the window, and $E_{\text{local}} \subset E_t$ includes the edges connecting those nodes.

% \subsubsection{Global Belief Graph}

% We construct the \emph{global belief graph} $G_{\text{global}} = (V_{\text{global}}, E_{\text{global}})$ by applying modularity-based clustering to the collision-free graph $G_t$ \cite{community2008}. The graph modularity $Q$ is used to identify densely connected clusters that serve as high-level nodes for global planning. It is formally defined as:

% \begin{equation}
% 	Q = \frac{1}{2m} \sum_{i,j} \left[ A_{ij} - \frac{k_i k_j}{2m} \right] \delta(c_i, c_j)
% 	\label{eq:modularity-def}
% \end{equation}

% Here, $m$ is the total number of edges in the graph, $A_{ij}$ is the $(i,j)$-th entry of the adjacency matrix, $k_i$ is the degree of node $i$, and $\delta(c_i, c_j)$ equals $1$ if nodes $i$ and $j$ are in the same community, and $0$ otherwise.

\subsubsection{Global Belief Graph with Modularity-Based Pruning}

We construct the \emph{global belief graph} $G_{\text{global}} = (V_{\text{global}}, E_{\text{global}})$ by jointly performing community detection and modularity-based pruning on the collision-free graph $G_t$. Instead of retaining all detected communities, we explicitly select a subset of structurally informative communities according to their modularity contribution, and only these communities are promoted as high-level nodes for global reasoning.

Specifically, we first apply modularity-based community detection on $G_t$ \cite{community2008}, where the graph modularity is defined as
\begin{equation}
Q = \frac{1}{2m} \sum_{i,j} \left[ A_{ij} - \frac{k_i k_j}{2m} \right] \delta(c_i, c_j),
\label{eq:modularity-def}
\end{equation}
with $m$ denoting the total number of edges, $A_{ij}$ the adjacency matrix, $k_i$ the degree of node $i$, and $\delta(\cdot)$ the community indicator.

Let $\mathcal{C}$ denote the set of communities detected from $G_t$. For simplicity, we ignore edge weights and directions and reorganize the modularity objective at the community level:
\begin{equation}
Q = \frac{1}{2m} \sum_{c \in \mathcal{C}} \sum_{i, j \in c}\left[ A_{ij} - \frac{k_i k_j}{2m} \right].
\label{eq:modularity-expanded}
\end{equation}
For each community $c$, let  $\sum{\mathrm{in}}$ denote the number of edges in $c$, and let $\sum{\mathrm{tot}}$ denote the number of edges connected to $c$. Thus, we have:
\begin{equation}
\sum_{i, j \in c} \frac{k_i k_j}{2m}
= \frac{(\sum{\mathrm{tot}})^2}{2m},
\label{eq:modularity-identity}
\end{equation}
the modularity can be rewritten as
\begin{equation}
Q = \frac{1}{2m} \sum_{c \in \mathcal{C}} \left[ \sum{\mathrm{in}} - \frac{(\sum{\mathrm{tot}})^2}{2m} \right].
\label{eq:modularity-simplified}
\end{equation}
Then, the modularity of community $c$ is
\begin{equation}
Q(c) = \sum{\mathrm{in}} - \frac{(\sum{\mathrm{tot}})^2}{2m}.
\label{eq:modularity-per-community}
\end{equation}
Rather than preserving all communities, we directly integrate a pruning step into the construction of $G_{\text{global}}$ by retaining only the top-$k$ communities with the highest modularity contributions:
\begin{equation}
\mathcal{C}' := \operatorname{argtopk}_{c \subseteq \mathcal{C}} Q(c).
\label{eq:modularity-prune}
\end{equation}

Each retained community $c \subseteq \mathcal{C}'$ is abstracted as a node in $V_{\text{global}}$, while edges in $E_{\text{global}}$ are induced by inter-community connectivity in $G_t$. In this way, the global belief graph is constructed directly from a pruned set of structurally coherent communities, yielding a compact yet informative high-level representation that significantly reduces reasoning complexity while preserving the dominant topological structure of the environment.

%%%%%%%%%%%%%%%%%%%%%%%%%%%%%%%%%%%%%%%%%%%%%%%%%%%%%%%%%%%%%%%%%%%%%%%%%%%%%%%%

\subsection{Slow-Thinking Module}

\subsubsection{Environment Conditioned Strategy Generation}
This part provides a principled mechanism for translating high-level environment descriptions into strategy-level guidance for autonomous exploration. Rather than directly prescribing actions or trajectories, it operates at the level of environment-conditioned reasoning, bridging semantic understanding and long-horizon planning objectives.

According to the intuitive impression formed after a brief exposure to the environment, we give a concise natural language description that summarizes the environment type and its composition. The slow-thinking module first performs structured environment characterization using an LLM. To ensure interpretability and robustness, the analysis is constrained to a predefined schema that decomposes the environment into three complementary aspects. \textbf{Spatial characteristics} capture the global layout and navigability of the environment, including openness, structural complexity, topological connectivity, and corridor width. \textbf{Obstacle characteristics} describe the distribution and structure of obstacles, such as their density, predictability, and vertical variation. \textbf{Exploration challenges} reflect task-level difficulties induced by the environment, including navigation difficulty, the likelihood of dead ends, and the necessity of backtracking.

\begin{tcolorbox}[fonttitle = \small\bfseries, title=Environment Characterization Example,colframe=gray!2!black,colback=gray!2!white,boxrule=1pt,boxsep=0pt,left=5pt,right=5pt,fontupper=\footnotesize, halign title = flush center]
	
	"{\color[RGB]{170, 193, 240}spatial characteristics}":
	\textbf{openness}: "confined",
	\textbf{complexity}: "moderate",
	\textbf{connectivity}: "low",
	\textbf{corridor width}: "narrow"
	
	"{\color[RGB]{213, 232, 212}obstacle characteristics}":
	\textbf{density}: "moderate",
	\textbf{predictability}: "irregular",
	\textbf{height variation}: "flat"
	
	"{\color[RGB]{248, 206, 204}exploration challenges}":
	\textbf{navigation difficulty}: "moderate",
	\textbf{dead end probability}: "moderate",
	\textbf{backtracking necessity}: "moderate"
\end{tcolorbox}

Based on the extracted environment characteristics, we instantiate an exploration strategy by populating a set of predefined strategy dimensions. These dimensions provide a structured representation of agent-level exploration behavior and are designed to capture recurring decision patterns across diverse environments. Specifically, the strategy is parameterized along four complementary axes: \textbf{spatial strategy}, \textbf{efficiency strategy}, \textbf{safety strategy}, and \textbf{task strategy}. Each axis governs a distinct aspect of exploration behavior, including coverage patterns and traversal order, energy and time trade-offs, risk sensitivity and obstacle handling, and completion objectives and information priorities. The environment characteristics directly condition the values assigned to these dimensions. For instance, environments exhibiting low topological connectivity increase tolerance for backtracking and enforce stronger awareness of escape routes, reflecting the higher risk of entrapment. Dense or irregular obstacles bias the strategy toward conservative motion, larger obstacle clearance, and cautious treatment of unexplored regions. Similarly, high navigation difficulty favors exploration behaviors that prioritize reliability and path quality over speed. In this way, environment semantics are systematically translated into structured exploration strategies that remain interpretable and explicitly grounded in environmental factors.

%Specifically, templates define exploration behavior along four dimensions: spatial strategy, which governs coverage patterns, directional bias, depth preference, and corridor handling; efficiency strategy, which regulates energy usage, time constraints, revisit policies, and tolerance for backtracking; safety strategy, which specifies obstacle clearance, risk sensitivity in unknown regions, dead-end handling, and escape-route awareness; and task strategy, which determines completion criteria, information priorities, and the trade-off between exploration speed and solution quality. By operating at this level of abstraction, the system avoids brittle, environment-specific heuristics while retaining strong inductive structure.

\begin{tcolorbox}[fonttitle = \small\bfseries, title=Strategy Example,colframe=gray!2!black,colback=gray!2!white,boxrule=1pt,boxsep=0pt,left=5pt,right=5pt,fontupper=\footnotesize, halign title = flush center]

	'\textbf{description}': 'Outdoor environment with natural obstacles and terrain variations',
	
	'\textbf{spatial}': {
		'coverage strategy': 'boundary first',
		'direction bias': 'perimeter following',
		'depth strategy': 'balanced depth breadth',
		'corridor handling': 'natural path'
	},
	
	'\textbf{efficiency}': {
		'energy policy': 'conservative',
		'time constraint': 'moderate',
		'backtrack tolerance': 'moderate',
		'revisit policy': 'avoid'
	},
	
	'\textbf{safety}': {
		'obstacle clearance': 'conservative',
		'unknown area approach': 'standard',
		'dead end handling': 'explore carefully',
		'escape route awareness': 'always maintain'
	},
	
	'\textbf{task}': {
		'completion criteria': 'time limited',
		'information priority': 'object detection',
		'quality vs speed': 'balanced'
	}
\end{tcolorbox}

\subsubsection{Graph Reasoning}

Given the pruned global belief graph $G_{\text{global}}$, a textual strategy prompt $x$ and the episode memory $m$, we perform iterative graph reasoning with the instance $\Pi$ of LLMs.  At reasoning depth $i$, the LLMs perform:

\begin{equation}
\tau_g = \Pi(G_{\text{global}},\, x, m)
\end{equation}

where $\tau_g=[v_{\mathrm{cur}},v_1,\dots,v_m]$ is a global path from the current node to an unexplored node.

%%%%%%%%%%%%%%%%%%%%%%%%%%%%%%%%%%%%%%%%%%%%%%%%%%%%%%%%%%%%%%%%%%%%%%%%%%%%%%%%

\subsection{Fast-Thinking Module}

\subsubsection{Policy Network}
The fast-thinking module operates on a structured observation that integrates the local graph, utility, and global path. At each time step, the observation is defined as $o_t=\left(G^*, \tau_t\right)$. 
Following~\cite{ariadne,cao2024deep,cao2025header}, we build an informative graph $G^*=\left(V_l^*, E_l\right)$, which shares the same edge set and node positions with the local graph $G_{\text{loacl}}$. In addition to the position $\left(x_i, y_i\right)$, each node $v_i^{\prime}=\left(x_i, y_i, u_i, g_i\right) \in V_l^*$ in the informative graph has two more properties: the utility $u_i$ and guidepost $g_i$ (a binary signal that denotes whether the location of the node is in the global paths). 
The planner selects a neighboring node $v_i \in \mathcal{N}\left(v_{\text {cur }}\right)$ as the next waypoint $w_t$, and the robot executes the action $a_t$ to move toward $w_t$.
Our policy network comprises attention-based encoder and decoder modules tailored to graph-structured inputs. We first compute query ($\mathbf{q}_i$), key ($\mathbf{k}_i$), and value ($\mathbf{v}_i$) vectors via learned linear transformations:
\begin{equation}
\mathbf{q}_i = W^q \mathbf{h}_i^{(q)},\quad
\mathbf{k}_i = W^k \mathbf{h}_i^{(k,v)},\quad
\mathbf{v}_i = W^v \mathbf{h}_i^{(k,v)},
\end{equation}
where $\mathbf{h}_i \in \mathbb{R}^{d_f}$ denotes the input node feature vector associated with node $v_i^{\prime}$, $W^q, W^k, W^v \in \mathbb{R}^{d_f \times d_f}$ are learnable projection matrices, and superscripts $(q)$ and $(k,v)$ indicate query or key/value projections, respectively. The scaled dot-product attention scores are then computed between pairs of nodes as $u_{ij} = \frac{\mathbf{q}_i^\top \mathbf{k}_j}{\sqrt{d_f}}$ and transformed into normalized attention weights subject to edge constraints indicated by an adjacency-based mask matrix $M$:

\begin{equation}
	\begin{aligned}
		w_{ij} &= 
		\frac{\exp(u_{ij}) \cdot (1 - M_{ij})}{\sum_{j'=1}^{n} \exp(u_{ij'}) \cdot (1 - M_{ij'})}, \\[4pt]
		M_{ij}&=\begin{cases}
			0, & (v_i,v_j)\in E \\[3pt]
			1, & (v_i,v_j)\notin E
		\end{cases}
	\end{aligned}
\end{equation}
where $E$ is the edge set defining the current adjacency relationships. Finally, the output node features $\mathbf{h}_i^{\prime} = \sum_{j=1}^{n} w_{ij}\mathbf{v}_j$ aggregates the weighted values across neighbors.

\subsubsection{Instruction Following}
\begin{figure}[thpb]
  \centering
  \includegraphics[scale=0.4]{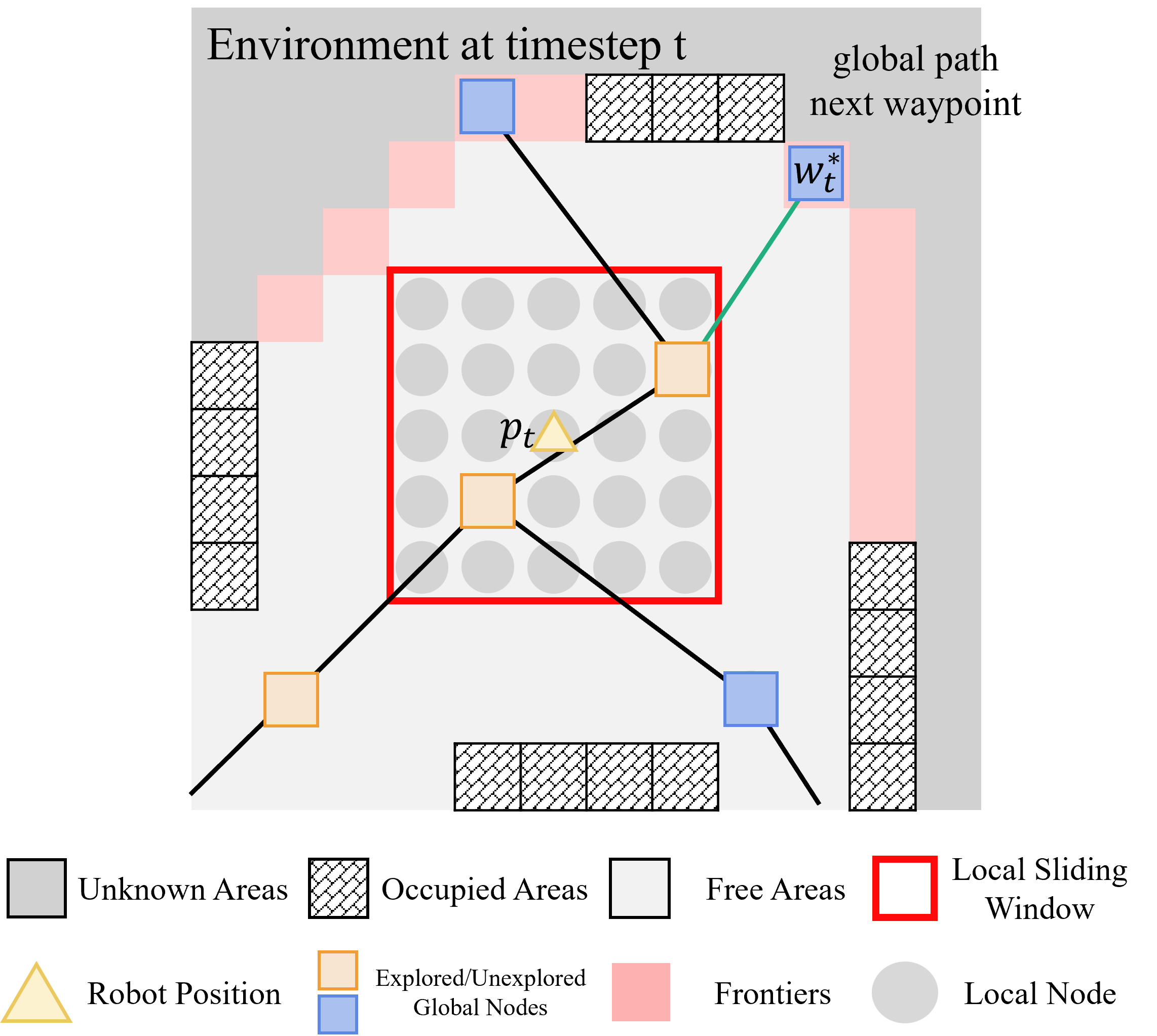}
  \caption{Illustration of the environment state at timestep $t$, showing the robot at position $p_t$, the policy-selected waypoint $w_t$, and the globally guided next waypoint $w_t^*$.}
  \label{obs}
\end{figure}

Building upon the reward design in~\cite{cao2025header}, we introduce an instruction following objective to encourage the fast-thinking policy to adhere to the long horizon guidance generated by the slow-thinking module. At decision step $t$, the fast-thinking policy is guided by a global path $\tau_t^*$, whose next waypoint is denoted as $w_t^*$, and selects a local waypoint $w_t$ for execution.

At a high level, following the global guidance reduces unnecessary detours and implicitly shortens the executed trajectory. While this objective can be expressed in terms of path length deviation, directly optimizing such quantities is impractical for reinforcement learning. Instead, we optimize a smooth surrogate that preserves the monotonic relationship with deviation from the global guidance. We define the normalized deviation between the policy-selected waypoint and the globally guided waypoint as
\begin{equation}
	d_t
	=
	\frac{\| w_t - w_t^* \|}
	{4 \, \Delta_{\text{node}} \sqrt{2}},
\end{equation}
where $\Delta_{\text{node}}$ denotes the node resolution. The normalization constant corresponds to an upper bound on admissible deviation measured in grid units, ensuring scale invariance across environments. The instruction following reward is then defined as an exponential penalty on the normalized deviation,
\begin{equation}
	r_t^{\text{dev}}
	=
	-
	\frac{e^{d_t} - 1}{e - 1},
\end{equation}
\begin{figure*}[t]
\centering
\subfloat[Indoor\label{fig:indoor}]{
    \includegraphics[width=0.32\textwidth]{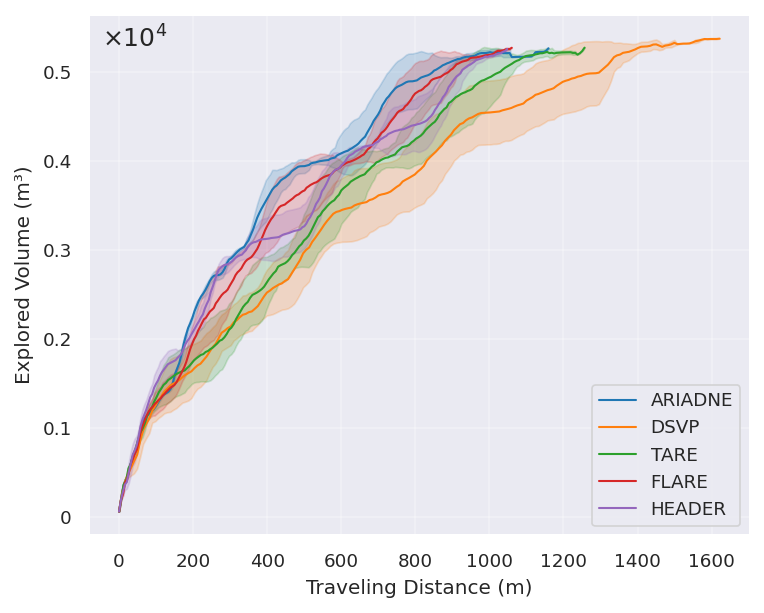}
}
% \hspace{0.02\textwidth}
\subfloat[Forest\label{fig:forest}]{
    \includegraphics[width=0.32\textwidth]{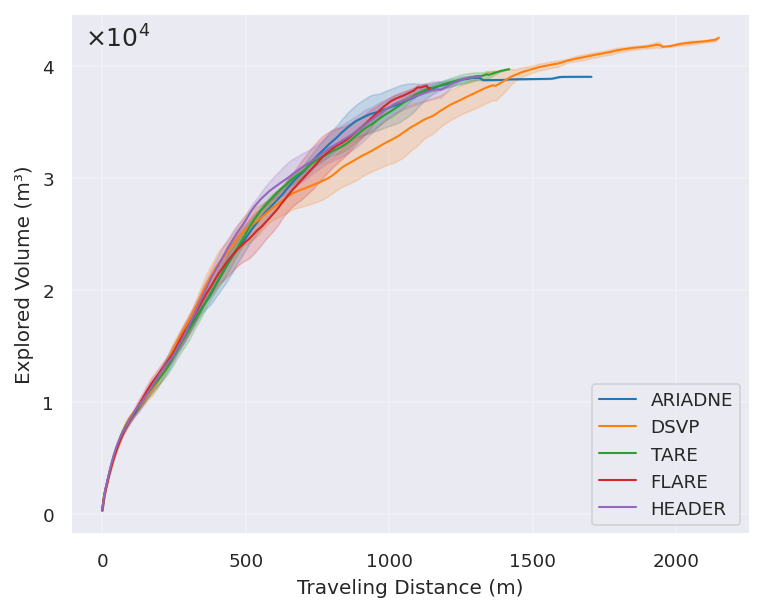}
}%
% \hspace{0.02\textwidth}
\subfloat[Warehouse\label{fig:warehouse}]{
    \includegraphics[width=0.32\textwidth]{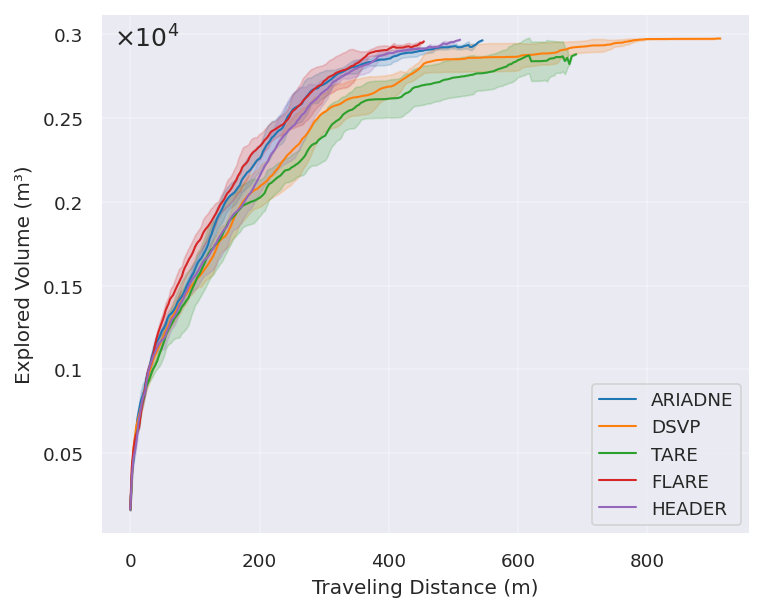}
}%
\caption{Performance comparison between FARE and conventional baselines across 10 runs per method in each environment.}
\label{fig:ros_results}
\end{figure*}

\begin{table*}[t]
\centering
\caption{\textbf{Comparison results in three environment.}}
\label{tab:results}

\begin{subtable}[t]{0.32\textwidth}
\centering
\caption{Indoor}
\begin{tabular}{c|cc}
\toprule
 & Distance ($m$) & Time ($s$)  \\
\midrule
DSVP & $1511 (\pm 75)$ & $931 (\pm 61)$ \\
TARE & $1209 (\pm 42)$ & $658 (\pm 23)$ \\
ARiADNE & $1053 (\pm 63)$ & $610 (\pm 24)$ \\
HEADER & $1030 (\pm 40)$ & $576 (\pm 26)$ \\
\midrule
\textbf{FARE} & $\textbf{1048} (\pm 13)$ & $\textbf{590} (\pm 10)$ \\
\bottomrule
\end{tabular}
\end{subtable}
\hfill
\begin{subtable}[t]{0.32\textwidth}
\centering
\caption{Forest}
\begin{tabular}{c|cc}
\toprule
 & Distance ($m$) & Time ($s$) \\
\midrule
DSVP & $2058 (\pm 92)$ & $1083 (\pm 60)$ \\
TARE & $1363 (\pm 43)$ & $711 (\pm 21)$ \\
ARiADNE & $1320 (\pm 81)$ & $790 (\pm 62)$ \\
HEADER & $1230 (\pm 72)$ & $725 (\pm 36)$ \\
\midrule
\textbf{FARE} & $\textbf{1090} (\pm 21)$ & $\textbf{680} (\pm 10)$ \\
\bottomrule
\end{tabular}
\end{subtable}
\hfill
\begin{subtable}[t]{0.32\textwidth}
\centering
\caption{Warehouse}
\begin{tabular}{c|cc}
\toprule
 & Distance ($m$) & Time ($s$) \\
\midrule
DSVP & $869 (\pm 42)$ & $582 (\pm 32)$ \\
TARE & $652 (\pm 31)$ & $366 (\pm 22)$ \\
ARiADNE & $521 (\pm 16)$ & $362 (\pm 40)$ \\
HEADER & $492 (\pm 17)$ & $286 (\pm 16)$ \\
\midrule
\textbf{FARE} & $\textbf{441} (\pm 15)$ & $\textbf{252} (\pm 8)$ \\
\bottomrule
\end{tabular}
\end{subtable}

\end{table*}

\begin{figure*}[t]
\centering
\subfloat[Indoor]{
    \includegraphics[height=4.5cm]{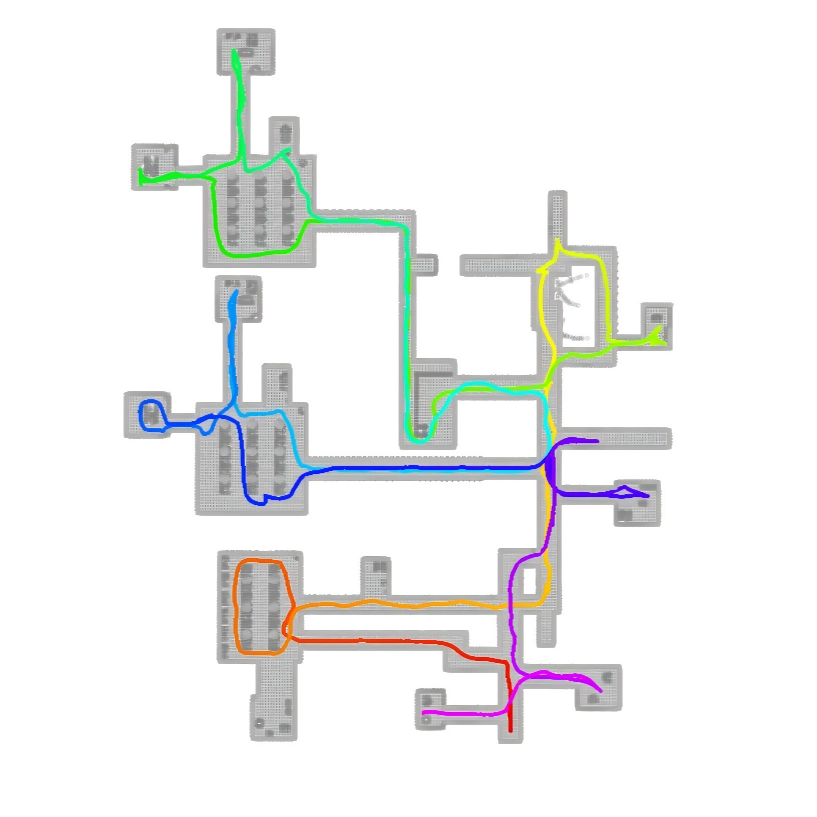}
}%
\hspace{0.06\textwidth}
\subfloat[Forest]{
    \includegraphics[height=4.5cm]{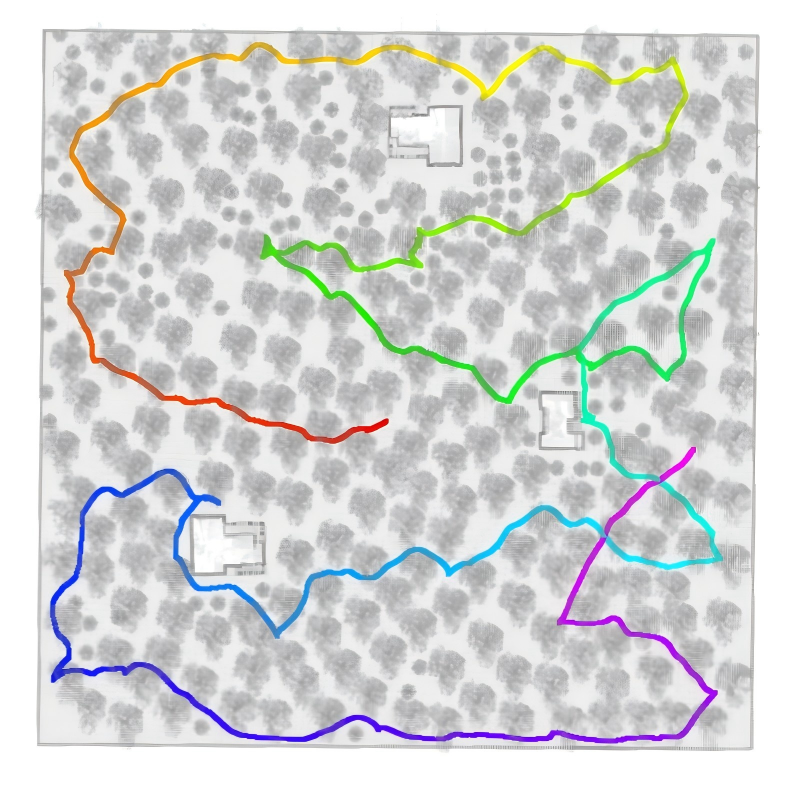}
}%
\hspace{0.07\textwidth}
\subfloat[Warehouse]{
    \includegraphics[height=4.5cm]{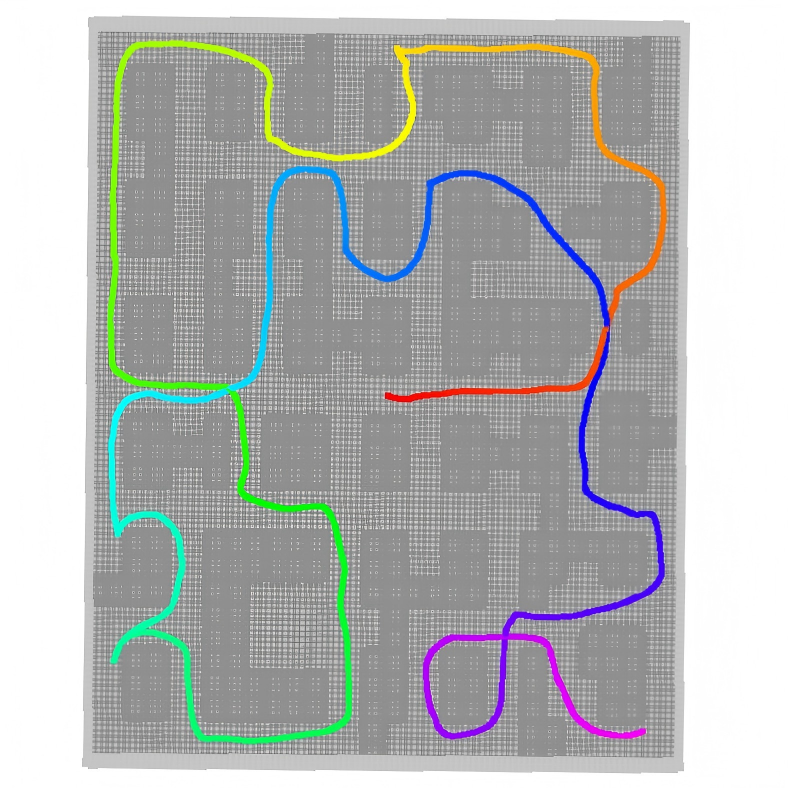}
}%
\caption{Demonstration of the exploration trajectories output by FARE in both indoor and outdoor Gazebo simulations. The Trajectory is color-coded to represent the robot's movement over time.}
\label{fig:ros_trj}
\end{figure*}
which yields $r_t^{\text{dev}} \in [-1,0]$. This formulation assigns mild penalties for small deviations while increasingly suppressing large deviations from the global guidance, providing smooth and stable gradients for policy optimization. By this construction, the waypoint deviation $d_t$ is monotonically related to detours from the global path. Therefore, minimizing the cumulative instruction-following penalty implicitly encourages shorter executed trajectories, while remaining fully compatible with efficient reinforcement learning.

%%%%%%%%%%%%%%%%%%%%%%%%%%%%%%%%%%%%%%%%%%%%%%%%%%%%%%%%%%%%%%%%%%%%%%%%%%%%%%%%
%%%%%%%%%%%%%%%%%%%%%%%%%%%%%%%%%%%%%%%%%%%%%%%%%%%%%%%%%%%%%%%%%%%%%%%%%%%%%%%%

\section{Experiments}

We conduct a set of experiments to evaluate the effectiveness and robustness of FARE. We first perform comparative evaluations in Gazebo simulation across three representative environments—\emph{indoor}, \emph{forest}, and \emph{warehouse}—to assess exploration performance under diverse structural characteristics. We then deploy FARE on a real mobile robot and validate its performance in a large-scale campus environment. All experiments use the same trained model and identical system configurations unless otherwise specified.

%%%%%%%%%%%%%%%%%%%%%%%%%%%%%%%%%%%%%%%%%%%%%%%%%%%%%%%%%%%%%%%%%%%%%%%%%%%%%%%%

\subsection{Comparison Analysis}
\begin{figure*}[t]
\centering
\subfloat[Agilex Scout-mini \label{fig:robot}]{
  \includegraphics[height=4cm]{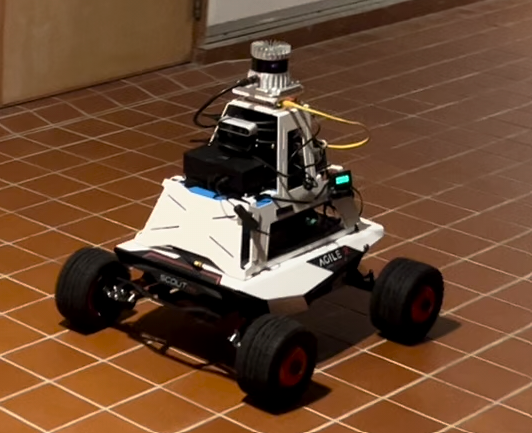}
}
\hspace{0.02\textwidth}
\subfloat[Indoor Teaching Building Map\label{fig:nusmap}]{
  \includegraphics[height=4cm]{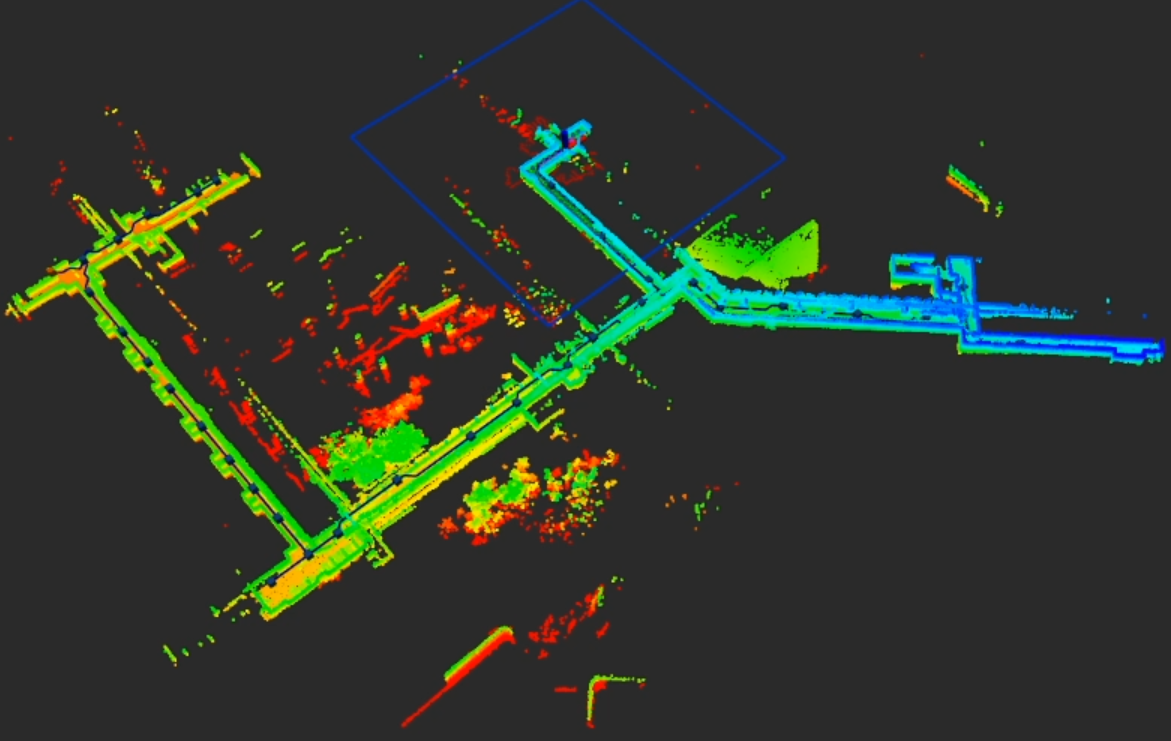}
}
\hspace{0.02\textwidth}
\subfloat[Exploration Trajectory\label{fig:nuspath}]{
  \includegraphics[height=4cm]{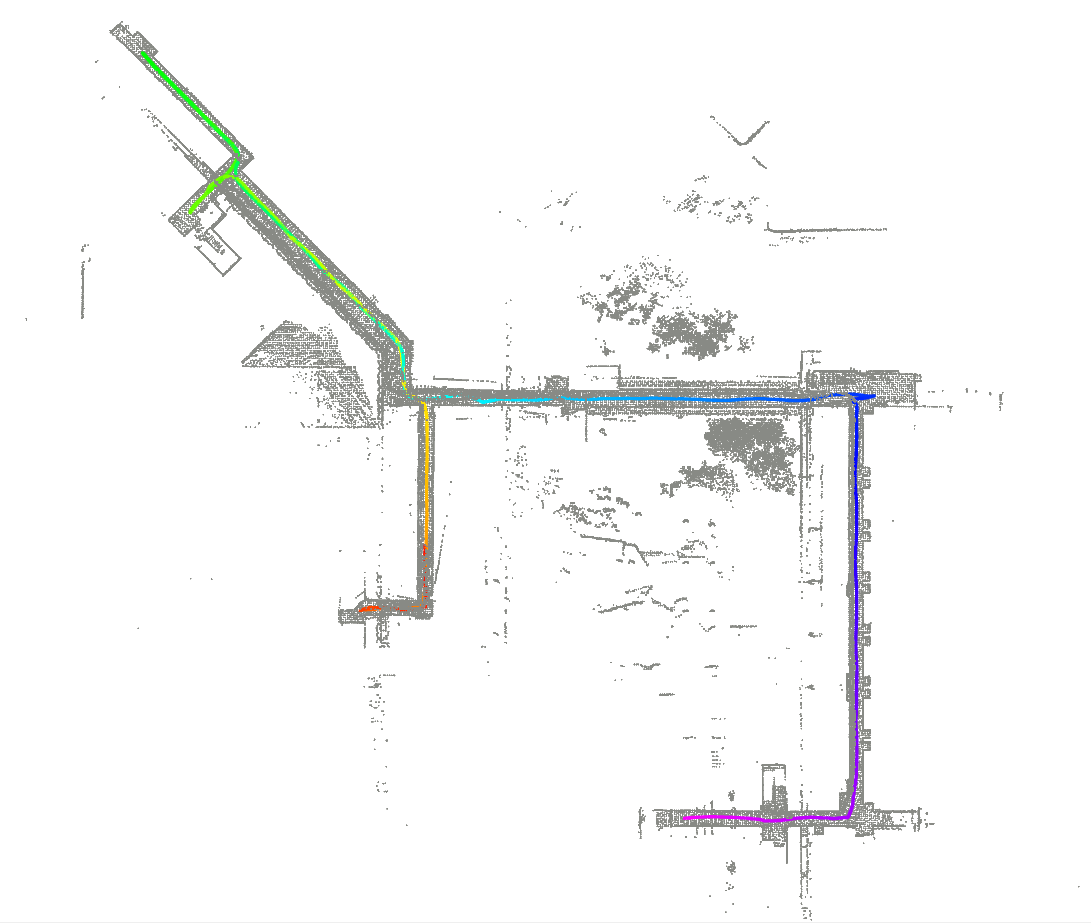}
}
\caption{Validations of FARE in real-world scenarios on a wheeled robot.}
\label{fig:harware}
\end{figure*}
We integrate FARE into the robot operating system (ROS) and compare it with representative state-of-the-art exploration planners, including \textbf{TARE}~\cite{cao2021tare}, \textbf{DSVP}~\cite{zhu2021dsvp}, \textbf{ARiADNE}~\cite{cao2024deep} and \textbf{HEADER}~\cite{cao2025header}. For a fair comparison, ARiADNE is implemented with the same graph rarefaction strategy as in~\cite{cao2024deep}, and all graph-related settings are kept consistent across methods. 
The comparison is conducted in Gazebo simulation across three benchmark environments: an \emph{indoor} environment, a \emph{forest} environment, and a \emph{warehouse} environment. All experiments are performed on a four-wheeled differential-drive robot equipped with a 16-channel 3D LiDAR, with a maximum speed of \(2\,\mathrm{m/s}\).

For different environments, FARE only adjusts the node resolution parameter \(\Delta_{\text{node}}\), which is varied from \(1.2\,\mathrm{m}\) to \(2.8\,\mathrm{m}\). The baseline planners generally require tuning of multiple parameters to achieve their best performance.
In addition, FARE is provided with a short environment description specific to each benchmark environment to condition the generation of the exploration strategy, with the slow-thinking module instantiated as \textbf{Qwen3-14B}. The environment descriptions are as follows: Indoor—modern indoor office building with long corridors, meeting rooms, and cubicle areas; Forest—outdoor forest environment with natural obstacles, trees, and uneven terrain; and Warehouse—indoor warehouse composed of densely arranged box stacks forming narrow aisles.
Each method is executed for 10 runs per environment. Travel distance and explored volume are reported in Fig.~\ref{fig:ros_results}, and quantitative results are summarized in Table~\ref{tab:results}.

FARE achieves performance on par with other baselines in the indoor environment, where the environment is compact and lacks distinctive global-level structure. In the forest environment, FARE achieves a clear reduction in travel distance and makespan, and this advantage further widens in the warehouse environment, where FARE obtains the shortest paths and fastest completion times among all methods. Specifically, as illustrated by the trajectory visualizations in Fig.~\ref{fig:ros_trj}, FARE does not rely solely on local frontier or utility signals. Instead, it systematically incorporates global structural cues during exploration. After reaching the boundary of the explored region, FARE tends to complete peripheral and corner areas early, rather than postponing them. In contrast, baseline methods often defer these regions and revisit them later, resulting in additional backtracking and reduced overall efficiency.

We attribute FARE’s performance gains to two key factors. First, our hierarchical design enables environment-adaptive planning, allowing FARE to adjust its global exploration strategy according to the environment character. Second, effective coordination between global guidance and local execution allows the agent to balance long-horizon objectives with local reactivity. The slow-thinking module provide global direction, while the fast-thinking module remain flexible to exploit nearby informative regions. 

%%%%%%%%%%%%%%%%%%%%%%%%%%%%%%%%%%%%%%%%%%%%%%%%%%%%%%%%%%%%%%%%%%%%%%%%%%%%%%%%

\subsection{Hardware Validation}

We validate \textsc{FARE} on an Agilex Scout-mini wheeled robot equipped with an onboard \textbf{Jetson AGX Orin}, which runs the LLM-based slow-thinking module using \textbf{Qwen3-14B}. An Ouster OS0-32 LiDAR is used for perception, and FastLIO2~\cite{xu2022fast} provides odometry and mapping. For all experiments, the maximum robot speed is set to \(1\,\mathrm{m/s}\), the sensor range to \(8\,\mathrm{m}\), and the fast-thinking module replanning frequency to \(1\,\mathrm{Hz}\).
Real-world validation is conducted in a $200m\times130m$ \emph{indoor teaching building} on campus. The environment consists of long corridors, rooms, and intersections, posing challenges in global reasoning and long-horizon exploration. We set the map resolution to \(\Delta_{\text{map}}=0.4\,\mathrm{m}\) and the node resolution to \(\Delta_{\text{node}}=0.8\,\mathrm{m}\). These parameters are selected prior to deployment and remain fixed throughout the experiment.
During deployment, \textsc{FARE} successfully explores the entire building without manual intervention. The system maintains stable runtime performance during global guidance generation and local policy execution. This hardware experiment demonstrates that \textsc{FARE} can effectively operate with onboard LLM inference and transfer from simulation to real-world environments.

%%%%%%%%%%%%%%%%%%%%%%%%%%%%%%%%%%%%%%%%%%%%%%%%%%%%%%%%%%%%%%%%%%%%%%%%%%%%%%%%
%%%%%%%%%%%%%%%%%%%%%%%%%%%%%%%%%%%%%%%%%%%%%%%%%%%%%%%%%%%%%%%%%%%%%%%%%%%%%%%%

\section{CONCLUSIONS}

In this work, we propose FARE, a hierarchical autonomous exploration framework that separates environment-conditioned global reasoning from fast local decision-making. FARE translates concise natural language environment descriptions into structured, interpretable exploration strategies, and performs LLM-based graph reasoning on a pruned global belief graph to generate adaptive global guidance. A fast-thinking policy integrates local graph structure, utility signals, and global paths, and is explicitly trained to follow long-horizon guidance while retaining local flexibility. Together, these components enable coherent exploration behaviors that reduce redundant backtracking and improve efficiency, as validated in both simulation and real-world experiments.

Future work will extend FARE to multi-robot exploration with explicit inter-agent coordination, and incorporate vision-based semantic perception to enable online detection of environment changes, better supporting hybrid environments with abrupt scene-type transitions. We also plan to investigate richer environment representations and three-dimensional action spaces to further enhance generality.

% A conclusion section is not required. Although a conclusion may review the main points of the paper, do not replicate the abstract as the conclusion. A conclusion might elaborate on the importance of the work or suggest applications and extensions. 


%%%%%%%%%%%%%%%%%%%%%%%%%%%%%%%%%%%%%%%%%%%%%%%%%%%%%%%%%%%%%%%%%%%%%%%%%%%%%%%%
%%%%%%%%%%%%%%%%%%%%%%%%%%%%%%%%%%%%%%%%%%%%%%%%%%%%%%%%%%%%%%%%%%%%%%%%%%%%%%%%

% \newpage
\bibliographystyle{IEEEtran}
\bibliography{IEEEexample}

\end{document}